\documentclass[10pt]{article}

\usepackage{arxiv}

\usepackage[utf8]{inputenc} 
\usepackage[T1]{fontenc}    
\usepackage{hyperref}       
\usepackage{url}            
\usepackage{booktabs}       
\usepackage{amsfonts}       
\usepackage{nicefrac}       
\usepackage{microtype}      
\usepackage{lipsum}
\usepackage{subcaption}
\usepackage{graphicx}
\usepackage{soulutf8}
\usepackage{color}

\title{Unsupervised Behavior Change Detection in Multidimensional Data Streams for Maritime Traffic Monitoring}

\author{
  Lucas May Petry\thanks{Corresponding author. Work developed during his visit to the Institute for Big Data Analytics at Dalhousie University.} \\
  Programa de Pós-Graduação em Ciências da Computação \\
  Universidade Federal de Santa Catarina (UFSC)\\
  Florianópolis, Brazil\\
  \texttt{lucas.petry@posgrad.ufsc.br} \\
  \And
  Amilcar Soares \\
  Institute for Big Data Analytics \\
  Dalhousie University \\
  Halifax, Canada \\
  \texttt{amilcar.soares@dal.ca} \\
  \And
  Vania Bogorny \\
  Programa de Pós-Graduação em Ciências da Computação \\
  Universidade Federal de Santa Catarina (UFSC)\\
  Florianópolis, Brazil\\
  \texttt{vania.bogorny@ufsc.br} \\
  \And
  Stan Matwin \\
  Institute for Big Data Analytics \\
  Dalhousie University \\
  Halifax, Canada \\
  \texttt{stan@cs.dal.ca} \\
}

\begin{document}
\maketitle



\section{Introduction}


The worldwide growth of maritime traffic and the development of the Automatic Identification System (AIS) has led to advances in monitoring systems for preventing vessel accidents and detecting illegal activities.
While preventing vessel accidents means saving money for shipping companies, it may also protect the marine fauna and flora from irreversible damage \cite{claramunt2017maritime}.
Furthermore, the integration of vessel traffic data with environmental and climatological data allows more complex analyses and a better understanding of the cause and effect of maritime events \cite{soares2019crisis}.
The overwhelming amount of data continuously generated by shipping companies, climate stations, and satellites, to name a few, poses challenges to maritime monitoring systems and their users.

Maritime traffic monitoring has been the subject of study of several works in the literature \cite{soares2019crisis,lane2010maritime,van2012abstracting,patroumpas2015event,lei2019mining}.
In particular, a few approaches have been proposed for the detection of events or behavior changes from AIS data, such as changes in the speed \cite{soares2019crisis,pitsikalis2019composite,wen2019semantic} or in the course of vessels \cite{pitsikalis2019composite,wen2019semantic,varlamis2019network}, proximity of vessels to the coast or to other vessels \cite{lei2019mining}, drifting or loitering behavior \cite{pitsikalis2019composite}, avoidance behavior \cite{lettich2016detecting}, illegal fishing \cite{patroumpas2015event} or possibly hazardous activity \cite{soares2019crisis}, among others.
Real-time event detection provides monitoring agents with interpretable semantic information about vessel behavior (e.g. assign to a vessel the behavior of illegal fishing activity for a certain period of time), allowing quick responses to events.
For example, detecting small vessels heading towards ice-infested waters and that are not equipped for handling this situation allows the decision maker to warn the captain in advance.
Such strategy avoids the deployment of a search and rescue mission, which might represent a high cost (e.g., lives, resources) to maritime authorities.
However, to the best of our knowledge, existing works are ad-hoc approaches, limited to detecting a restricted set of predefined vessel behaviors.
Such methods not only prevent the detection of unforeseen events but also require the assistance of domain specialists for defining rules and thresholds that characterize each event.

In this work, we describe research gaps and challenges in machine learning for vessel behavior change and event detection, considering several constraints imposed by real-time data streams and the maritime monitoring domain.
As a starting point, we investigate how unsupervised and semi-supervised change detection methods may be employed for identifying shifts in vessel behavior, aiming to detect and label unusual events.

\section{Research Challenges}

We describe the challenges of maritime traffic monitoring with respect to three different aspects: multidimensional streaming data, detection of behavior changes, and the extraction and storage of knowledge acquired from the data for posterior reasoning.

\begin{figure}[!t]
    \centering
    \begin{subfigure}{0.27\textwidth}
        \centering
        \includegraphics[width=\linewidth]{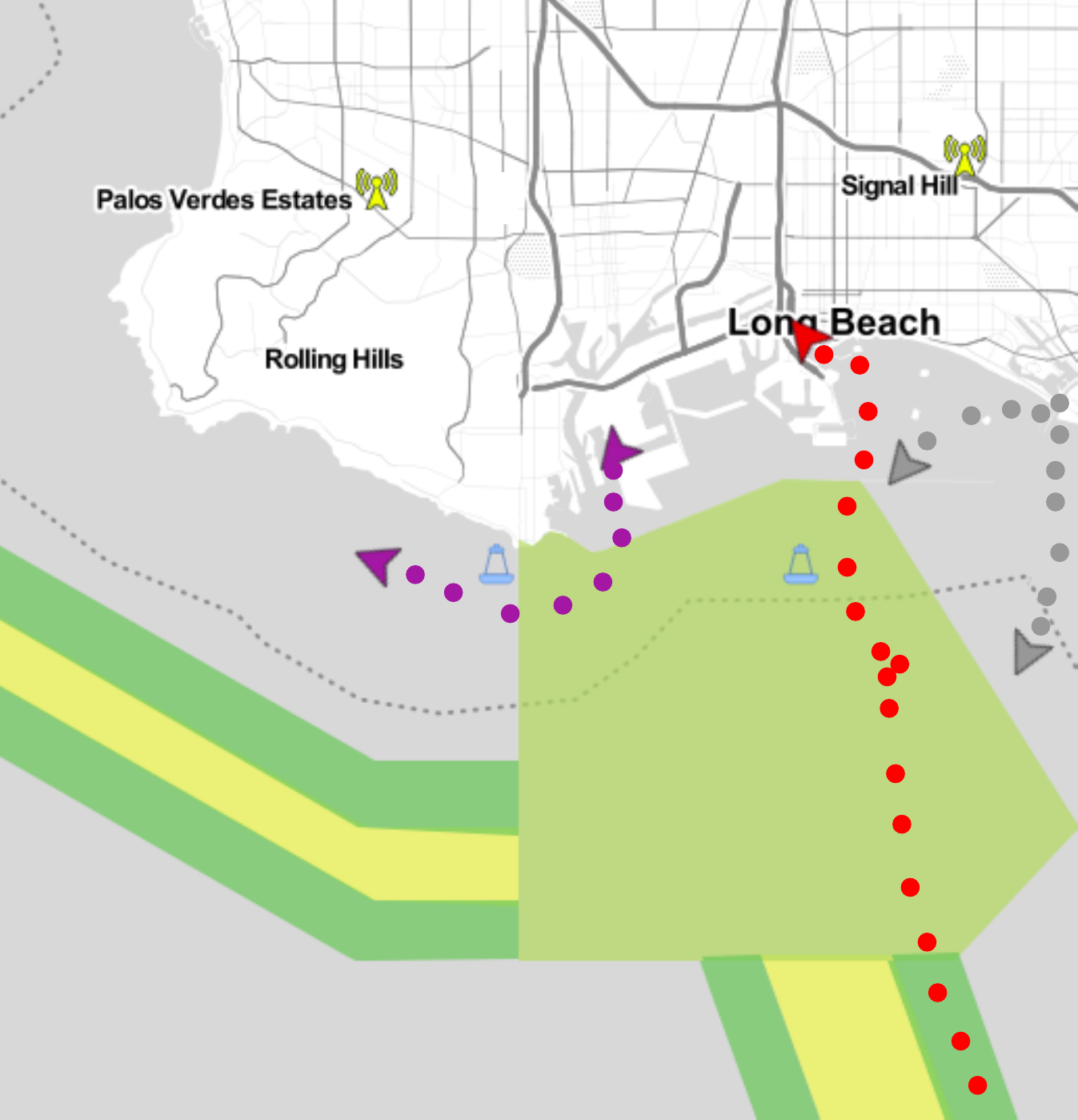}
    \end{subfigure}
    \hspace{8pt}
    \begin{subfigure}{0.65\textwidth}
        \centering
        \includegraphics[width=\textwidth]{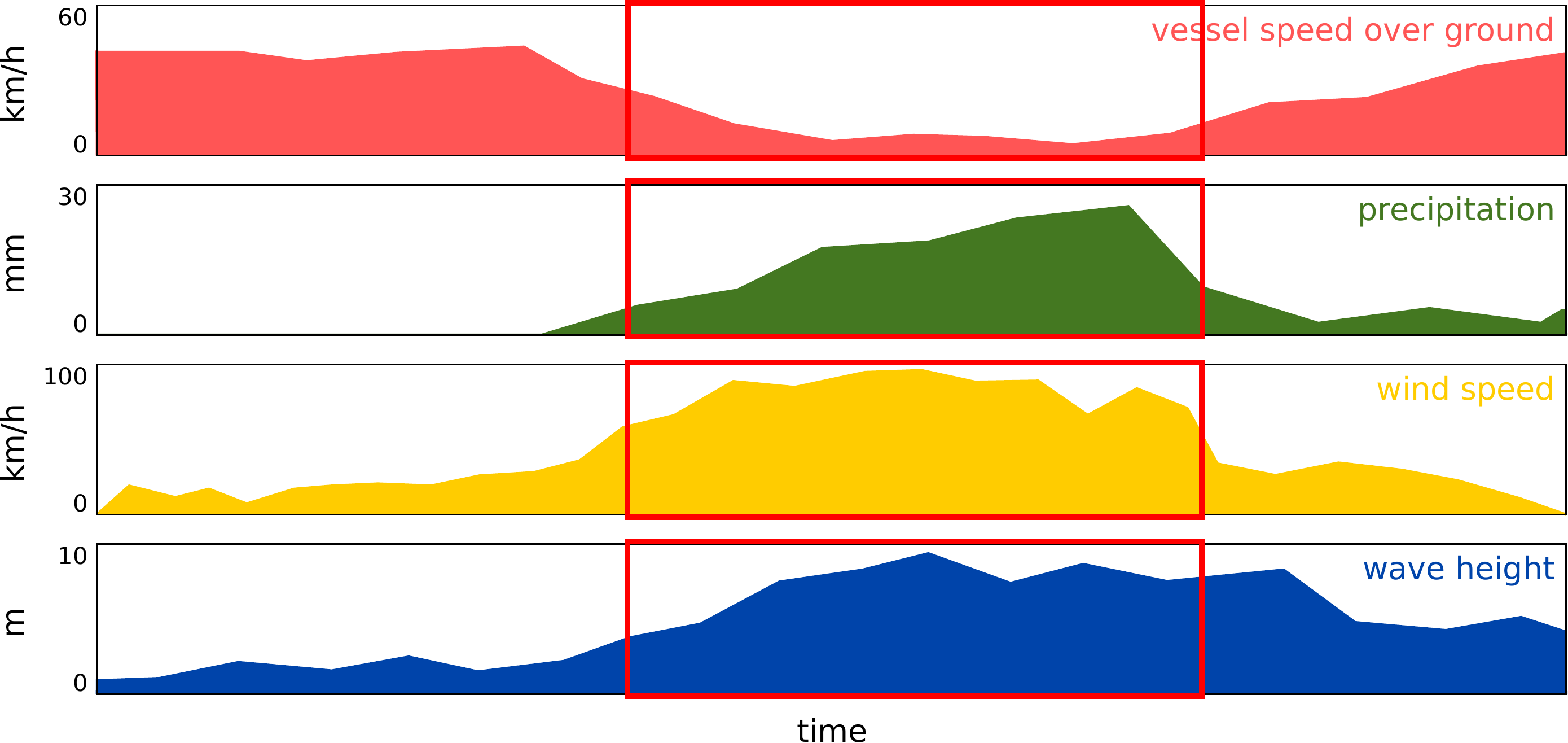}
    \end{subfigure}
    \caption{AIS data, climate stations, marine buoys, and traffic lanes visualization example (left). Toy streaming data from different sensors with highlighted hazardous event of a tropical storm (right).}
    \label{fig:ais_example}
    \vspace{-15pt}
\end{figure}

\textbf{Multidimensional data streams.}
As highlighted in \cite{soares2019crisis}, combining AIS data with other streaming data (e.g., ocean and climate conditions) is a challenge to current maritime monitoring systems. 
Figure~\ref{fig:ais_example}~(left) shows a visual example of vessel trajectory data with climate stations and marine buoys that collect climate and ocean data, respectively.
Streaming AIS data is often sparse (i.e., Satellite AIS), with minutes or even hours before the arrival of new data points.
Besides, sensors from multiple data sources have different sampling rates, and their spatial location may be static or dynamic.
Integrating such data for analysis is a non-trivial task, with temporal and spatial constraints.
Lastly, integrated streaming data from different sensors provide algorithms with a high number of new features, thus suffering from the curse of dimensionality.
Handling data sparsity and their high dimensionality with real-time constraints can be challenging for machine learning algorithms.

\textbf{Behavior change detection.}
Most of the existing works for vessel traffic monitoring do not learn from the data.
Instead, they propose specific algorithms with behavior rules and thresholds defined beforehand for detecting a restricted set of vessel activities.
Defining such rules requires the knowledge from domain specialists, thus introducing a lot of human bias in the analysis.
Moreover, defining strict thresholds (e.g., vessel speed above a specified threshold configures high-speed) can be difficult and may induce the identification of irrelevant behavior.
Hence, the ability to learn behavior patterns from vessel and sensor data can alleviate the burden on the user for setting up specific event rules.
Another issue with existing approaches is that they mainly analyze the individual behavior of vessels.
However, some anomalies may only be detected when analyzing the collective behavior of vessels.
For instance, a vessel switching off the AIS device on purpose to hide some illegal activity can be detected by analyzing the AIS messages captured in an area where several others were transmitting at that time.

\textbf{Knowledge extraction and storage.}
To enable user agents to make data-driven decisions about suspicious or dangerous vessel activity, discovered behavioral patterns must be interpretable.
Interpretability is, perhaps, the main advantage of existing works, since events related to vessel behavior are explicitly defined by rules.
On the other hand, a learning algorithm must be able to identify behavior patterns explainable to the user.
This not only enables the understanding of the user about newly discovered vessel activities but also allows knowledge storage for detecting recurring behavior.
Detecting recurring behavior may be essential, especially if validated with the user, as it might be something the user wants to keep monitoring or is worth investigating.

\section{Preliminary Work and Research Directions}

We are currently investigating and exploiting existing change detection methods, and concept drift approaches for analyzing vessel behavior change.
In streaming data, concept drift is commonly referred to as the detection of significant changes in data distribution \cite{gonccalves2014comparative}.
Figure~\ref{fig:ais_example}~(right) illustrates a maritime scenario with AIS (e.g., vessel speed over ground), environmental (e.g., precipitation and wind speed), and ocean data (e.g., wave height).
We highlight in Figure~\ref{fig:ais_example}~(right) the time window (in red) that corresponds to a vessel going through a deteriorating event, such as a tropical storm.
The event is characterized by the increase of precipitation, wind speed, and wave height, as well as by the decrease of the vessel speed over ground.

Several works for change detection in data streams were designed to work along with a supervised learning model, detecting drifts based on the error rate of the learner \cite{bach2008paired,ross2012exponentially,sethi2016grid}.
However, due to the high cost and difficulty of manually identifying and labeling vessel behaviors, these methods are not suitable for detecting changes in the maritime domain.
Although unsupervised approaches have been proposed for detecting concept drift, they are either limited to univariate data \cite{bifet2007learning} or simply correlate data batches of features individually \cite{lee2012detection}.
Other works based on Principal Component Analysis (PCA) have been proposed for unsupervised change detection in multidimensional data streams \cite{kuncheva2013pca,qahtan2015pca}.
Even though they treat the high dimensionality of data, they do not extract interpretable behavior patterns of the changes found, nor do they provide means for detecting recurring events, i.e., similar behavior changes that happen over time.

More recently, Hallac et al. \cite{hallac2017toeplitz} proposed the Toeplitz Inverse Covariance-based Clustering (TICC) method for segmenting multivariate sensor data into sequences of states or clusters (i.e. behavior patterns).
TICC represents each of these states as a Markov Random Field (RMF), which provides information of direct dependencies between variables and, therefore, exhibits a certain level of interpretability about each state.
TICC is able to detect recurring behavior by assigning similar segments of data to the same cluster.
However, the number of clusters is fixed and must be defined by the user, meaning that the number of different behaviors present in the data should be known a priori.
Additionally, TICC is not directly suitable for streaming data, as it assumes that all data is available at the same time.
We are presently studying and testing this approach on our data for detecting behavior changes in the maritime domain.
We expect to find good results, being able to further improve TICC to the multidimensional data streaming scenario, considering the challenges aforementioned.
Lastly, we want to consider feedback from the user when detecting vessel behavior activity, so that the algorithm may learn whether or not a particular behavior is interesting to the user and should be monitored.
Visual Interactive Labeling (VIL) \cite{bernard2017unified} and Active Learning (AL) \cite{junior2017analytic} are two strategies that might help to assist the user in the process of annotating the data or reducing the number of training samples, respectively.

\section*{Acknowledgments}

This work was partially supported by Global Affairs Canada (GAC) via the Emerging Leaders in the Americas Program (ELAP).

\bibliographystyle{unsrt}  
\bibliography{references}

\end{document}